\RequirePackage[hyphens]{url}
\documentclass[sigconf]{acmart}
\usepackage{booktabs} 
\usepackage{multirow}
\usepackage{flushend}

\setcopyright{rightsretained}

\begin{document}
	\title{Exploring the use of Time-Dependent Cross-Network Information for Personalized Recommendations}
	
	\author{Dilruk Perera}
	\affiliation{%
		\institution{School of Computing}
		\institution{National
			University of Singapore}
	}
	\email{dilruk@comp.nus.edu.sg}
	
	\author{Roger Zimmermann}
	\affiliation{%
		\institution{School of Computing}
		\institution{National
			University of Singapore}
	}
	\email{rogerz@comp.nus.edu.sg}
	

	\begin{abstract}
		The overwhelming volume and complexity of information in online applications make recommendation essential for users to find information of interest. However, two major limitations that coexist in real world applications (1) incomplete user profiles, and (2) the dynamic nature of user preferences continue to degrade recommender quality in aspects such as timeliness, accuracy, diversity and novelty.  To address both the above limitations in a single solution, we propose a novel cross-network time aware recommender solution. The solution first learns historical user models in the target network by aggregating user preferences from multiple source networks. Second, user level time aware latent factors are learnt to develop current user models from the historical models and conduct timely recommendations. We illustrate our solution by using auxiliary information from the Twitter source network to improve recommendations for the YouTube target network. Experiments conducted using multiple time aware and cross-network baselines under different time granularities show that the proposed solution achieves superior performance in terms of accuracy, novelty and diversity.
	\end{abstract}

	\copyrightyear{2017} 
	\acmYear{2017} 
	\setcopyright{acmlicensed}
	\acmConference{MM '17}{October 23--27, 2017}{Mountain View, CA, USA}
	\acmPrice{15.00}
	\acmDOI{10.1145/3123266.3123447}
	\acmISBN{978-1-4503-4906-2/17/10}

	\ccsdesc[500]{Information systems~Recommender systems}
	\ccsdesc[500]{Information systems~Social recommendation}
	\ccsdesc[500]{Information systems~Collaborative filtering}
	
	\keywords{Recommender system; Cross-network; Time aware; User profiling}

	\maketitle
	
	\section{Introduction}
\label{intro}
Recommender systems are essential tools for the success of Online Social Networks (OSNs) such as Twitter, YouTube and Facebook as they provide mutual benefits for both users and OSNs. For example, Netflix reports that at least 75\% of Netflix user interactions are initiated by their internal recommender engine\footnote{\url{http://techblog.netflix.com/2012/04/netflix-recommendations-beyond-5-stars.html}}. Therefore, over the past few decades, constant efforts have been made to improve the quality of personalized recommendations under different aspects such as timeliness, accuracy, diversity and novelty \cite{mclaughlin2004collaborative}. However, we identified two major limitations that continue to coexist in practical recommender systems and degrade overall recommender quality.
\par
(1) \textbf{Incomplete user profiles} -- Typical recommender systems depend only on the limited information available within a single OSN to create user profiles for both new and existing users. However, the notorious data scarcity issues in a single OSN have significantly hindered the development of rich user profiles with diverse user preferences \cite{tan2014cross}. For existing users, the very high item to user ratio in an OSN limits their interactions to a small subset of available items. Hence, the resulting sparse interactions are often insufficient to effectively comprehend their preferences (\textit{data sparsity problem}) \cite{sang2015cross,hurley2013personalised}. For new users, the absence of any historical interactions makes it infeasible to gauge their preferences (\textit{cold start problem}). Accordingly, the resulting incomplete user profiles limit the achievable recommender quality in terms of accuracy, diversity and novelty.
\par
(2) \textbf{The dynamic nature of user preferences} -- User preferences toward items constantly change due to factors such as redefinitions of user inclinations, the emergence of new items, and popularity variations of items \cite{koren2010collaborative, camposperformance}. Therefore, recommender systems should constantly update user profiles to incorporate user interest drifts, and thereby develop timely user models for effective recommendations. Otherwise, even the most complete user profiles become obsolete, and recommender quality would decline over time \cite{wu2016unfolding}. Furthermore, user preference changes can be comparatively rapid in certain applications while moderate in others. For example, users in short video applications (e.g., YouTube and Vine) tend to have abrupt changes in their video preferences, whereas in a movie streaming application (e.g., Netflix), user preferences tend to be fairly constant. Hence, for certain applications, the integration of user preference dynamics is an even greater need to achieve a reasonable recommender quality.
\par
The extant literature contains distinct efforts to address both coexisting limitations separately. However, these limitations bound the achievable recommender quality in different aspects. Therefore, we propose a consolidated solution to address both limitations for all types of users (i.e., new and existing). To create comprehensive user profiles by mitigating data scarcity issues, we integrate user interaction data from multiple OSNs. According to the Global Web Index for 2015, a typical user maintains around 5 different social media accounts\footnote{\url{https://www.globalwebindex.net/blog/internet-users-have-average-of-5-social-media-accounts}}. For example, a user that shares posts and interacts with friends on Facebook, can upload and share photos on Flickr, and watch favorite TV shows on YouTube. Furthermore, we hypothesize that the specialization of OSNs for different activities (e.g., YouTube for entertainment related activities and Twitter for news related activities) motivates users to maintain a \textquotedblleft multiple presence\textquotedblright \space in the digital world, and user activities performed on these OSNs represent user interests from multiple perspectives. Exploiting information from multiple OSNs leads to an increase in accuracy by mitigating data scarcity issues, and the diversity of user preferences exploited leads to novel and diverse item recommendations for users. Furthermore, in addition to the typical user-item context, we incorporate a temporal context to capture the timestamps of user-item interactions on these networks. Consequently, by aggregating historical user preferences across networks and exploiting their changes over time, we learn a more effective representation of current user preferences to predict future user interactions.
\par
Thus, the proposed time aware cross-network recommender solution transfers auxiliary user interaction information from source networks to a target network and provides recommendations for target network items. We demonstrate the effectiveness of our solution using auxiliary information from the Twitter source network for video recommendations on the target YouTube network. The proposed solution is general and not limited to these networks; therefore, it can easily be extended to incorporate multiple source networks to achieve higher recommendation quality.  
\par
In this paper, first, we conduct a data analysis to validate the motivation and feasibility of our solution. Second, we present the proposed recommender solution that utilizes time-stamped, cross-network information for both new and existing user recommendations. Third, we test our solution in multiple experimental settings and compare the results in terms of accuracy, diversity and novelty, against various baseline methods under varying time granularities. 
\par
We summarize the main contributions of this paper as follows:
\begin{itemize}
	\setlength\itemsep{0em}
	\item To the best of our knowledge, this is the first attempt to develop a model based, time aware cross-network recommender solution.
	\item We created a new dataset with Twitter and YouTube interactions along with their corresponding timestamps. 
	\item We conducted a data analysis to demonstrate that Twitter is a good auxiliary information source for YouTube recommendations, YouTube users change their preferences within the examined time frame, and social networks are biased toward different topics (network biases).
	\item The effectiveness of the proposed solution for conducting timely recommendations is demonstrated using multiple time granularities under several experiments.
\end{itemize}

\section{Literature Review}
\label{litrev}
We reviewed existing recommender solutions under two separate, but related streams of research as follows:
\par 
\textbf{Cross-network Recommendation:} The use of information across multiple OSNs makes personalized recommendations comparatively more robust against data sparsity and cold start problems \cite{mehta2005ontologically}. For example, Delicious achieved a 10\% increase in recommender precision using supplementary information from Flicker and Twitter \cite{abel2011analyzing}. Researchers have further explored various combinations of source and target networks such as Google+ for YouTube \cite{deng2013personalized}, Twitter for YouTube \cite{roy2012socialtransfer} and Wikipedia for Twitter \cite{osborne2012bieber} to enhance recommender quality. These approaches have used a variety of information available on these networks such as social relationships (e.g., contacts and follows) and behavioral data (e.g., tags and tweets) \cite{yan2013friend, sang2015cross}. Despite successful attempts to enrich user profiles with auxiliary information, they do not capture user preference dynamics, which limits the achievable recommender quality.
\par
\textbf{Time aware Recommendation:} Existing time aware Collaborative Filtering (CF) approaches are mainly categorized as time aware memory-based and model-based methods. Memory-based methods capture similarities between users and/or items for recommendations and adopt multiple approaches to weight most recent interactions compared to least recent ones \cite{lathia2009temporal,ding2005time}. In contrast, model-based methods mainly use data mining and machine learning techniques to find underlying patterns from user-item interaction data. The recommender systems domain generally prefers model-based methods due to advantages such as robustness against data sparsity, scalability, and improved prediction accuracy. The winning attempt in Netflix Prize contest showed the significant quality improvements achievable with time aware model-based methods compared to memory-based methods \cite{koren2010collaborative}. Various time aware solutions proposed in the literature are based on techniques such as Dynamic Matrix Factorization (DMF) \cite{wang2016recommending}, Linear Dynamical Systems (LDS) \cite{chua2013modeling}, Probabilistic MF (PMF) \cite{salakhutdinov2007probabilistic} and Bayesian MF (BMF) \cite{salakhutdinov2008bayesian}. However, such solutions typically assume gradual evolutions in user preferences and only consider the relationships between consecutive time intervals in a single network. In contrast, we exploit users\textquoteright \space historical preferences from multiple OSNs and learn their underlying complex relationships to infer current preferences in a target network. 
\par
While the time aware and cross-network combination is yet to be exploited, the solution proposed by Deng et al. \cite{deng2015twitter} comes closest to our attempt. Their memory-based YouTube video recommender solution integrates users\textquoteright \space long-term YouTube preferences with short-term preferences in Twitter hot topics. However, the proposed approach is limited in the following ways: A memory-based method is inherently limited by data sparsity, scalability and low accuracy; preferences extracted from YouTube are treated as static; fails to provide diverse recommendations since users are assumed to be interested in a single topic at any given time; short term preferences are extracted only based on interactions with current hot topics, hence interactions with all other topics are ignored; and do not provide recommendations for new users.
\par
Therefore, considering the above limitations, we developed the first model-based time aware cross-network recommender solution, which aggregates user preferences from multiple networks and considers their drifts over time to conduct high quality and timely recommendations.

\section{Data Analysis}
\label{dtaann}

\subsection{Dataset}
\label{dtaset}
Due to the lack of publicly available cross-network datasets with time-stamped user interactions, we created a new dataset as follows: First, we extracted users with both Twitter and YouTube interactions from two cross-network datasets \cite{lim2015mytweet,yan2014mining}. Second, user IDs were used to scrape time-stamped user interaction data that spanned over a 12 month period from 1\textsuperscript{st} March 2015 to 29\textsuperscript{th} February 2016. Specifically, we downloaded user tweets as Twitter interactions and metadata of  videos either liked or added to playlists by users as YouTube interactions. The resultant dataset contained 14,133 users with 12,148,994 tweets and 254,659 YouTube video interactions.

\subsection{Feasibility}
\label{feasab}
We conducted an initial data analysis to validate the motivation and feasibility of our solution, and also to justify the choice of source (Twitter) and target (YouTube) networks by answering two main questions: (1) is Twitter a good auxiliary information source for YouTube recommendations? and (2) do YouTube users change their preferences within the examined time frame?
\par
To answer the above two questions, we pre-processed the dataset as follows: First, we used a topic modeling approach to project user interactions on both Twitter and YouTube to a cross-network topical space. This allowed for a direct comparison between heterogeneous interactions from both networks (e.g., tweets and liked videos). We assumed that a single tweet relates to a single topic and each user\textquoteright s tweet collection formed a \textit{document}. Similarly, each YouTube video was considered a single document and the associated texts (e.g., titles and descriptions) as \textit{words}. The resultant documents from both networks formed the corpus, and the Twitter-Latent Dirichlet Allocation (Twitter-LDA) \cite{zhao2011comparing} was used for topic modeling because of its effectiveness in processing short and noisy tweets. Consequently, the topics associated with each item (i.e., tweet and YouTube video) were identified. Then, by considering user interactions with these items and cosponsoring timestamps, each user was represented by a collection of topical distributions in historical time intervals. These distributions were referred to as \textit{absolute topical distributions}, where each frequency value determined the user preference level towards a topic in the corresponding time interval. Hence, for a given user $u_i$, considering his source network interactions, the absolute topical distributions were denoted by $Sa_i = \{\boldsymbol{sa_i^1};...;\boldsymbol{sa_i^t}\} \in \mathbb{R}^{T\times K^t}$ where, $T$ is the number of time intervals and $K^t$ is the number of topics. Each vector $\boldsymbol{sa_i^t} = \{sa_i^{(t,1)},..., sa_i^{(t,K^t)}\} \in \mathbb{R}^{1\times K^t}$ represents the source network topical distribution at time interval $t$, and $sa_i^{(t,k)}$ represents the frequency of the $k^{th}$ topic. Similarly, $Ta_i \in \mathbb{R}^{T \times K^t}$  represents the absolute topical distributions in the target network. Finally, we normalized each time-wise topical distribution ($\boldsymbol{sa_i^t}$ and $\boldsymbol{ta_i^t}$) using the total frequency of each distribution ($\Sigma_{k=1}^{K^t} sa_i^{(t,k)}$ and $\Sigma_{k=1}^{K^t} ta_i^{(t,k)}$). Consequently, we obtained normalized vectors $\boldsymbol{\overline{sa}_i^t} \in \mathbb{R}^{1 \times K^t}$ and $\boldsymbol{\overline{ta}_i^t} \in \mathbb{R}^{1 \times K^t}$ for source and target networks respectively for fair comparisons across networks.
\par
To answer the first question: Is Twitter a good auxiliary information source for YouTube recommendations, we examined whether YouTube users express similar preferences on Twitter at the same time. Hence, we calculated histogram intersections between normalized topical distributions on both networks in corresponding time intervals ($\boldsymbol{\overline{sa}_i^t}$ and $\boldsymbol{\overline{ta}_i^t}$ ) and obtained time-wise user preference overlaps on both networks. Subsequently, for each user, the average percentage of YouTube preferences expressed on Twitter at corresponding time intervals ($O_{yt}$) was calculated as follows:
\begin{equation}
O_{yt_i} = \frac{1}{T} \times \sum_{t=1}^{T}\Bigg(\frac{\Sigma_{k=1}^{K^t} min(\overline{sa}_i^{(t,k)}, \overline{ta}_i^{(t,k)})}{\Sigma_{k=1}^{K^t} \overline{ta}_i^{(t,k)}}\Bigg)
\end{equation}

\begin{figure}
	\includegraphics[width=\linewidth]{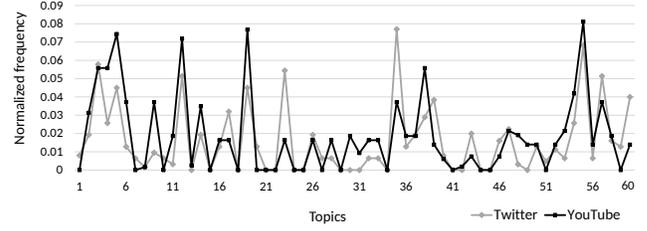}
	\caption{An example user's YouTube and Twitter topical distributions within the same time interval shows similar interests on both networks.}
	\label{fig:pics1_intersect}
\end{figure}

We observed that on average, users express 26\% of their YouTube interests on Twitter, and for 80\% of users, this value was as high as 43\% (see Figure \ref{fig:pics1_intersect} for a sample user). For example, a sample user (see Figure 1) shows that he is similarly interested in a \textquotedblleft social media related \textquotedblright \space topic (\#3) and a \textquotedblleft sports related\textquotedblright \space topic (\#37) in both networks at a given time interval. Therefore, we concluded that Twitter is a good auxiliary information source for YouTube recommendations.
\par
To answer the second question: Do YouTube users change their preferences within the examined time frame, we considered the intersection between normalized YouTube distributions in consecutive time intervals ($\boldsymbol{\overline{ta}_i^t}$ and $\boldsymbol{\overline{ta}_i^{t+1}}$). We observed that the average intersection was as low as 12\%, which implied that user preferences tend to vary considerably across time intervals and is therefore vital to incorporate these changes in a recommender solution. Note that the analysis was conducted with varying time intervals (i.e., biweekly and monthly) and the results were averaged.

\subsection{Network Biases}
\label{netbia}
In addition to the above analysis, we conducted an exploratory study to demonstrate that different networks are biased toward different topics. We incorporated this network property in the model development stage (see Section \ref{moddev}). We justified the network biases as follows: First, we calculated the network level topical distributions for source (i.e., for all $N$ users, $\Sigma_{i=1}^N \Sigma_{t=1}^T \boldsymbol{sa_i^t}$) and target (i.e., for all $N$ users, $\Sigma_{i=1}^N \Sigma_{t=1}^T \boldsymbol{ta_i^t}$) networks. Second, we normalized these distributions by their corresponding total frequency values ($\Sigma_{i=1}^N \Sigma_{t=1}^T \Sigma_{k=1}^{K^t} sa_i^{(t,k)}$ and $\Sigma_{i=1}^N \Sigma_{t=1}^T \Sigma_{k=1}^{K^t} ta_i^{(t,k)}$ for source and target networks). Finally, for each topic, we compared corresponding normalized frequency values on both networks and determined the relative network biases toward topics.
\par
\begin{figure}[t]
	\includegraphics[width=\linewidth]{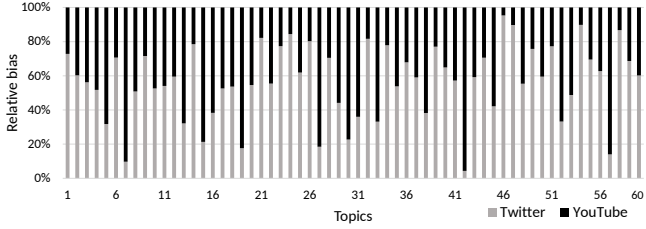}
	\caption{A comparison of normalized network level topical frequencies for Twitter and YouTube shows network biases toward different topics.}
	\label{fig:pics2_normalized_both}
\end{figure}

We conducted the experiment with varying number of topics (i.e., 5-100) and found that on average, 21\% of topics are 80\% or more frequent (biased) on one network than the other, and that 43\% of topics are 70\% or more frequent on one network than the other (see Figure \ref{fig:pics2_normalized_both} for experimental results for 60 topics). Upon further exploration of topics, it was evident that YouTube was more biased towards entertainment related topics (e.g., movie and video related) compared to Twitter, which was highly biased towards informative topics (e.g., events and news related). 

\begin{figure}[t]
	\includegraphics[width=\linewidth]{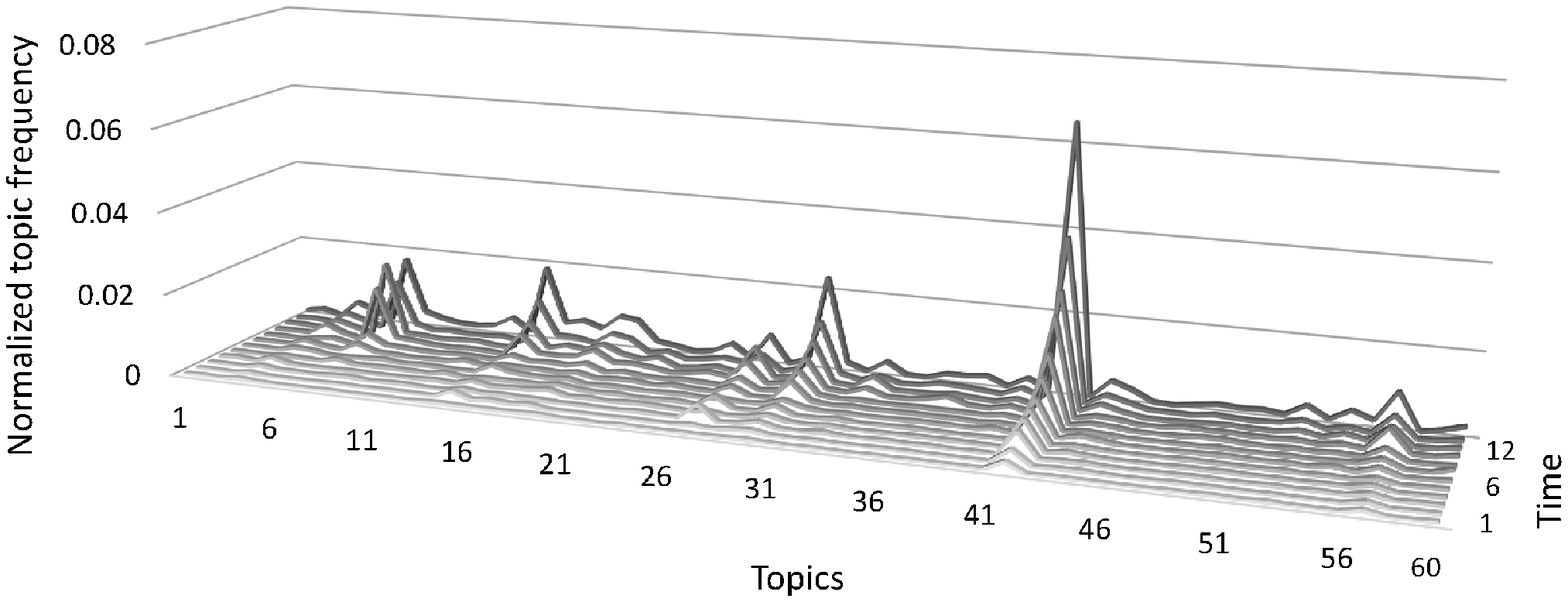}
	\caption{Normalized YouTube topical distributions over a 12 month period shows the consistency of topical biases on the network.}
	\label{fig:pics3_normalized_yt}
\end{figure}
\begin{figure}[t]
	\includegraphics[width=\linewidth]{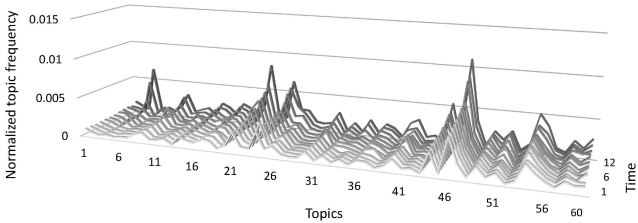}
	\caption{Normalized Twitter topical distributions over a 12 month period shows the consistency of topical biases on the network.}
	\label{fig:pics4_normalized_tw}
\end{figure}

\par
Furthermore, we observed such topical biases to be consistently present throughout the 12 month period (see Figures \ref{fig:pics3_normalized_yt} and \ref{fig:pics4_normalized_tw}). Hence, it was evident that both networks are significantly biased toward different topics, and it is an inherent property of the network.

\section{Model Preliminaries}
\subsection{Problem Formulation}
\label{modpre}
We denoted each existing YouTube user $u_e \in U_e = [Tw_{u_e}^t,$ \ $Yt_{u_e}^t]$ at time interval $t$ using tweets $Tw_{u_e}^t$ and video interactions $Yt_{u_e}^t$ that span over $T=\{1,...,t\}$ different time intervals. Each new YouTube user $u_n \in U_n = [Tw_{u_n}^t]$ at time interval $t$ was denoted only using a tweet collection as they were yet to interact with YouTube videos. Each tweet  $tw_{u_i} \in Tw_{u_i}^t$  and video $yt_{u_i} \in Yt_{u_i}^t$  contained textual tweet contents and textual video metadata (e.g., title and description) respectively along with their timestamps.
\par
Given the set of users $U= U_n \cup U_e$ and their user-video preference matrix $R^t$ at time interval $t$ (see Section \ref{prefmat}), we formulated video recommendation as a time aware Top-N recommendation task. For each user $u_i$ at time interval $t$, we predicted a set of $N$ items that the user would most likely interact with, in near future time intervals $(t+1,...)$. Hence, similar to the specification by Campos et al. \cite{camposperformance}, the recommendation task is formally specified as follows:
\begin{equation}
\begin{aligned}
\forall u_i \in U, t \in T, V_N^{*}(u_i,t) &= \bigcup_{j=1}^{N}v_j^*(u_i,t) : v_j^*(u_i, t) \\ &= \operatorname*{arg\,max}_{v \in V - V_{j-1}^*(u_i, t)} P(u_i, v, t)
\end{aligned}
\end{equation}

\noindent where, the preference function $P \colon U \times V \times T \rightarrow R^t$ and $V$ denotes the set of videos.

\subsection{Matrix Factorization}
\label{matfac}
Matrix Factorization (MF) \cite{koren2009matrix} is one of the most effective CF techniques, motivated by its ability to handle the high dimensionality and sparseness in typical user-item rating matrices. MF assumes user preferences can be modeled using multiple low rank latent factors, which are learnt from user rating patterns with minimal information loss to describe the original rating matrix. In its basic form, MF maps both users and items to a latent factor space of $K$ dimensionality where, the inner product between factors are used to model user-item interactions. Specifically, the $i^{th}$ user and the $j^{th}$ item are represented as latent vectors $\boldsymbol{u_i} \in \mathbb{R}^{K \times 1}$ and $\boldsymbol{v_j} \in \mathbb{R}^{K \times 1}$ where, the elements in $\boldsymbol{v_j}$ represent the level of association between the item and factors, and the elements in $\boldsymbol{u_i}$ represent the level of user interest towards items with high values for corresponding factors \cite{koren2009matrix}. Hence, the resulting dot product $\boldsymbol{u_i}^T\boldsymbol{v_j} \in \mathbb{R}$ captures the $i^{th}$ user\textquoteright s interest (user rating) towards the $j^{th}$ item. 
The user and item vectors are learnt by minimizing the regularized squared error between the approximated and observed values during the training phase. The corresponding loss function $L$ is as follows:

\begin{equation}
L = \operatorname*{min}_{\boldsymbol{u_*},\boldsymbol{v_*}} \sum_{(i,j) \in \varepsilon} (r_{ij} - \boldsymbol{u_i}^T \boldsymbol{v_j})^2 + \lambda(\parallel \boldsymbol{u_i} \parallel^2 + \parallel \boldsymbol{v_j} \parallel^2)
\end{equation}
where, $\varepsilon$ represents the set of observed ratings, and $r_{ij}$  is the explicit rating given by the $i^{th}$  user to the $j^{th}$ item. The regularization term $(\parallel\boldsymbol{u_i} \parallel^2+ \parallel \boldsymbol{v_j} \parallel^2)$ is added to prevent over fitting of user and item vectors, and the $\lambda$ parameter is experimentally set to control the degree of regularization.

\section{Proposed Model}
\label{propmod}
\subsection{The Preference Matrix ($\boldsymbol{R^t}$) }
\label{prefmat}
The standard user interaction matrix ($R$) does not incorporate timing information of user ratings (i.e., it represents user preferences uniformly throughout the entire interaction history), and therefore fails to provide a good measure of users\/' current preferences. For example, consider two users who have given similar ratings for a set of Football related videos; one is a continuous follower, while the other had developed a sudden interest only during last years FIFA world cup. The standard $R$ matrix does not capture the timing of the user preference change. Hence, the matrix encodes both users as having similar current preferences, which leads to outdated recommendations for the second user.  Therefore, we replaced the standard $R$ matrix in MF with a time aware user-video preference matrix $(R^t)$, which accounts for user preference changes over time and represents users\/' current/retained preferences.
\par
Our $R^t$ matrix is based on the following three assumptions: (1) similar to the standard $R$ matrix, user preferences are captured by their interactions (i.e., videos liked or added to playlists), (2) current preferences can be approximated using historical preferences, and (3) user preferences decay over time, hence the most recent preferences have a higher impact on current preferences than the least recent ones. Therefore, we utilized a time based decaying function to represent user preferences at time interval $t$, which is denoted by $R^t \in \mathbb{R}^{N \times M}$ for $N$ number of users and $M$ number of videos. Each element $r_{ij}^t = \Sigma_{c=1}^{C_{ij}} e^{\beta \hat{t}_c} \in R^t$ represents the $i^{th}$ user\textquoteright s preference for the $j^{th}$ item at time interval $t$ where, $C_{ij}$ denotes the number of interactions with the item, $\beta$ is a weighting parameter, and $\hat{t}_c \in \{1,...,t\}$ denotes the time interval index of the corresponding interaction $c$. Note that the $r_{ij}^t$ values are positively correlated with the frequency $(C_{ij})$ and recency $(t_c)$ of interactions. Furthermore, if the $i^{th}$  user has not interacted with the $j^{th}$ item (i.e., $c = 0$), $r_{ij}^t$ is initially set to 0 and the proposed model estimates the preference value to determine possible future interactions. Despite the increase in constant time operations (e.g., data access and arithmetic), the inclusion of timing information in $R^t$ does not increase the overall time complexity. Both $R$ and $R^t$ calculations take $O(J)$ time where $J$ is the number of interactions.
\subsection{Model Development}
\label{moddev}
Absolute topical distributions (see Section \ref{feasab}) are an imprecise representation of true user preferences as they are affected by two main network level factors: (1) network trends (e.g., US presidential election) and (2) network biases toward different topics (see Section \ref{netbia}). These factors influence users to interact more with trendy and biased items despite their actual preference levels; hence, the absolute topical frequency values related to such items get elevated and results in obscure representations. Therefore, to obtain an accurate representation of user preferences, we calculated their absolute frequency values relative to the corresponding state of the network, and thereby offset the effects from network level factors. The resulting representation is named as the \textit{relative topical distributions} of each user. 
\par
For each user $u_i$, the time-wise relative topical distributions in the source network are denoted by $Sr_i = \{\boldsymbol{sr_i^1}; ...; \boldsymbol{sr_i^t}\} \in \mathbb{R}^{T \times K^t}$ where, $\boldsymbol{sr_i^t} = \left\{\frac{sa_i^{(t, 1)}}{s_*^{(t, 1)}}, ..., \frac{sa_i^{(t,  K^t)}}{s_*^{(t, K^t)}}\right\} \in \mathbb{R}^{1 \times K^t}$ represents the relative topical distribution at time interval $t$, $sa_i^{(t, k)}$ denotes the absolute topical frequency value for the $k^{th}$ topic at time interval $t$, and $s_*^{(t,k)}= \Sigma_{i=1}^N sa_i^{(t,k)}$ indicates the total frequency for the $k^{th}$ topic across all users at time interval $t$. Similarly, $Tr_i \in \mathbb{R}^{T \times K^t}$ represents the relative topical distributions in the target network. 
\par
The relative topical distributions mitigate data heterogeneity by mapping user interactions from source and target networks to a homogeneous cross-network topical space. Then, we used two transfer functions to map network level preferences from the topical space to a target network user space, which allowed them to be used for recommendations in the target network. We used the transfer functions $M_S \in \mathbb{R}^{K^t \times K}$ and  $M_T \in \mathbb{R}^{K^t \times K}$ for the source and target networks respectively where, $K$ is the number of latent features in the target network user space. Thus, each user $u_i \in U_e$ is profiled in the target network user space as a collection of transferred historical  preferences from both source ($Us_i = Sr_i \cdot M_S \in \mathbb{R}^{T \times K}$) and target ($Ut_i = Tr_i \cdot M_T \in \mathbb{R} ^ {T \times K}$) networks. 
\par
Furthermore, we assumed users\/' current/retained preferences can be approximated using their historical user preferences (see Section \ref{prefmat}). Hence, for each user $u_i$, we learnt a time aware aggregation vector $\boldsymbol{t_i} \in \mathbb{R}^{1 \times T}$ to compute and combine the contributions of historical transferred preferences (i.e., $Us_i$ and $Ut_i$) and effectively model current preferences in the target network. 
\par
\begin{figure}[t]
	\includegraphics[width=\linewidth]{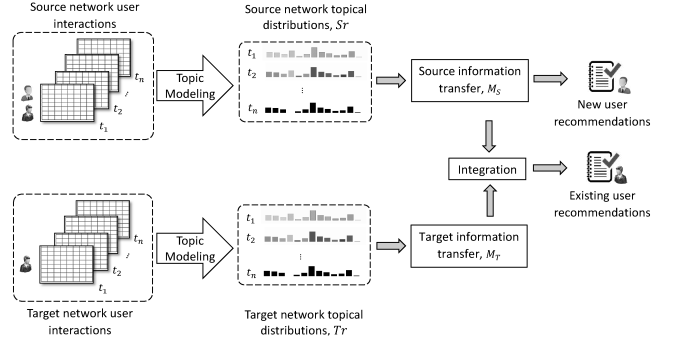}
	\caption{Overview of the proposed solution.}
	\label{fig:model}
\end{figure}

Figure \ref{fig:model} showcase the high level overview of the proposed solution. Transferred user preferences from the source network are directly used for new user recommendations. However, for existing user recommendations, transferred preferences from both source and target networks are integrated together.
\par
Thus, the corresponding loss function to optimize existing user predictions at current time interval $t$ is as follows:

\begin{equation}
\begin{split}
L = \operatorname*{min}_{\boldsymbol{t_*},M_S,M_T,\boldsymbol{v_*}} \sum_{(i,j) \in \varepsilon} \Big(r_{ij}^t - \boldsymbol{t_i} (Sr_i \cdot M_S + D_i \cdot Tr_i \cdot M_T)  \boldsymbol{v_j}\Big)^2 \\ + \lambda (||\boldsymbol{t_i}||^2 + ||M_S||^2 + ||M_T||^2 + || \boldsymbol{v_j}||^2)
\end{split}
\raisetag{2\normalbaselineskip}
\end{equation}

\noindent where, $r^t_{ij}$ is the user preference value of the $i^{th}$ user towards the $j^{th}$ item at current time interval $t$, $\boldsymbol{v_j} \in \mathbb{R}^{K \times 1}$ is the latent representation of the $j^{th}$ item, and the time-wise diagonal matrix for the $i^{th}$ user $D_i \in \mathbb{R}^{T \times T}$  weights the contributions from the transferred target network information for the current user preferences at different time intervals. Each element $D_i^{\hat{t}} = \gamma |T_{u_i}^{\hat{t}}|/|S_{u_i}^{\hat{t}}|$ weights the contribution at time interval $\hat{t}$ where, $\gamma$ is a weighting parameter, $|T_{u_i}^{\hat{t}}|$ and $|S_{u_i}^{\hat{t}}|$ represent the number of target and source network interactions of the user at time interval $\hat{t}$.
\par
We utilized the popular Stochastic Gradient Descent (SGD) technique to minimize the loss function and update the parameters for each observation in $\varepsilon$ using a learning rate $\mu$ as follows:
\begin{equation}\boldsymbol{t_i} \leftarrow \boldsymbol{t_i} + 2\mu\Big( e_{ij}  \boldsymbol{v_j}^T ({M_S}^T \cdot {Sr_i}^T + {M_T}^T \cdot {Tr_i}^T \cdot D_i) - \lambda \boldsymbol{t_i}\Big)\end{equation}
\begin{equation}M_S \leftarrow M_S + 2\mu\Big(e_{ij} ({Sr_i}^T \cdot \boldsymbol{t_i}^T \cdot \boldsymbol{v_j}^T) - \lambda M_S\Big)\end{equation} 
\begin{equation}M_T \leftarrow M_T + 2\mu \Big(e_{ij} ({Tr_i}^T \cdot  D_i \cdot \boldsymbol{t_i}^T \cdot \boldsymbol{v_j}^T) - \lambda M_T\Big)\end{equation} 
\begin{equation}\boldsymbol{v_j} \leftarrow \boldsymbol{v_j} + 2\mu \Big(e_{ij} ({M_S}^T \cdot {Sr_i}^T + {M_T}^T \cdot {Tr_i}^T \cdot D_i)\boldsymbol{t_i}^T - \lambda \boldsymbol{v_j}\Big)\end{equation}
where, $e_{ij} = r_{ij}^t - \boldsymbol{t_i}(Sr_i \cdot M_S + D_i \cdot Tr_i \cdot M_T)\boldsymbol{v_j}$. Once we learnt the parameters for each user $u_i \in U_e$, their current preference values for all items are obtained as follows: 
\begin{equation}\dot{r}_{i*}^t = \boldsymbol{t_i}(Sr_i \cdot M_S + D_i \cdot Tr_i \cdot M_T)V\end{equation} 

Subsequently, the model utilized the parameters learnt to conduct new user recommendations using the transferred historical source network user preferences. Since both new and existing user recommendations were conducted using the information within the same time period, we assumed that the matrices learnt (i.e., $M_S$ and $V$) are time consistent, and therefore suitable for new user recommendations. However, since target network interactions are not available for new users (i.e., $R^t$ is empty), the time aware vectors (i.e., $\boldsymbol{t_i}$) cannot be learnt for each individual user. Therefore, an approximated value is calculated based on the average vector values learnt for existing users. Hence, the time aware aggregation vector for new users is calculated as $\boldsymbol{t_{newi}} = \frac{1}{N_{ex}} \Sigma_{i=1}^{N_{ex}} \boldsymbol{t_i}$ where, $N_{ex}$ denotes the number of existing users.

Finally, for each new user $u_i \in U_n$, current preference values for all items are obtained as follows: 
\begin{equation}\ddot{r}_{i*}^t = \boldsymbol{t_{newi}} \cdot Sr_i \cdot M_S \cdot V\end{equation} 

\section{Experiments}
\label{experim}
The absence of standardized testing protocols in recommender systems literature has led to biased evaluation methods. For example, temporal overlaps between training and testing datasets provide undue prediction advantages that do not exist in the real world \cite{camposperformance}. Therefore, to avoid such biases, we divided training and testing datasets using a time-wise threshold. User interactions preceding the threshold were considered historical (training set) and the remaining were set as future interactions (testing set) for predictions. Hence, $\forall r_{tn}, r_{te}, t(r_{tn}) < t(r_{te})$ where, $r_{tn} \in Tn$, and $r_{te} \in Te; Tn, Te$ and $t(x)$ represent the training and testing datasets, and the timestamp of rating $x$, respectively. 

\subsection{Experimental Setup}
\label{expset}
We demonstrate the effectiveness of our model using data from Twitter (source) and YouTube (target) networks. To create an experimental dataset with adequate user interactions, we filtered out users with less than 5 interactions on each network, videos with less than 2 user interactions and tweets that were not explicitly labeled as English. The resultant dataset comprised of 3,134 users with 13,238 YouTube videos and 2,998,229 tweets, and the sparsity of the user-video matrix was 99.67\%. Then, all users were sorted by their YouTube interaction counts in ascending order and split evenly into two user groups (new and existing) where users with less interaction counts were assigned to the new user group. 
\par 
We designed four experimental settings (see Table \ref{tab:expsett}) by dividing the dataset using two time aware granularities (biweekly and monthly intervals), and two training and testing ratios (83/17 and 92/8). Note that only existing users have both historical and future YouTube interactions and for new users, historical YouTube interactions were hidden. 
\begin{table}[h]
	\small
	\centering
	\caption{Model evaluation dataset settings.}
	\label{tab:expsett}
	\begin{tabular}{|l|l|l|l|}
		\hline
		Experiment & Granularity & Training set & Testing set \\ \hline
		Exp. 1 & \multirow{2}{*}{Biweekly} & \begin{tabular}[c]{@{}l@{}}First\\   40 weeks (83\%)\end{tabular} & \begin{tabular}[c]{@{}l@{}}Last 8\\   weeks (17\%)\end{tabular} \\ \cline{1-1} \cline{3-4} 
		Exp. 2 &  & \begin{tabular}[c]{@{}l@{}}First\\   44 weeks (92\%)\end{tabular} & \begin{tabular}[c]{@{}l@{}}Last 4\\   weeks (8\%)\end{tabular} \\ \hline
		Exp. 3 & \multirow{2}{*}{Monthly} & \begin{tabular}[c]{@{}l@{}}First\\   10 months (83\%)\end{tabular} & \begin{tabular}[c]{@{}l@{}}Last 2\\   months (17\%)\end{tabular} \\ \cline{1-1} \cline{3-4} 
		Exp. 4 &  & \begin{tabular}[c]{@{}l@{}}First\\   11 months (92\%)\end{tabular} & \begin{tabular}[c]{@{}l@{}}Last 1\\   month (8\%)\end{tabular} \\ \hline
	\end{tabular}
\end{table}

To evaluate the proposed model, we implemented three time based (TimePop, Time-Biased KNN and TimeMF) and two cross-network based (Unified and ACNRS) baselines as follows: 
\begin{itemize}
	\setlength\itemsep{0em}
	\item \textbf{TimePop}: Calculates the most popular N items for the latest time interval (i.e., last two weeks or month) and recommends these items to all users. 
	\item \textbf{TBKNN}: The Time-Biased KNN method by Campos et al. \cite{campos2010simple} calculates $K$ neighbors for each user and uses their recent interactions to predict the target user interactions.  We used multiple $K$ values varying from 4 to 50 as per the original experiments and results were averaged for comparisons.
	\item \textbf{TimeMF}: Based on the standard MF technique (see Section \ref{matfac}). However, to incorporate the temporal context into MF, instead of the typical $R$ matrix, uses the $R^t$ matrix introduced in our method (see Section \ref{prefmat}).
	\item \textbf{Unified}: Similar to our experiments, the cross-network Unified method \cite{yan2015unified} is demonstrated for YouTube recommendations using Twitter auxiliary information. Hence, the Unified baseline facilitates direct comparison of performance when utilizing timing information in the cross-network domain.
	\item \textbf{ACNRS}: Alternative cross-network recommender system is a variation of our solution, which does not model user interest drifts over time. ACNRS uses the same set of factors as in the proposed solution. However, it uses the standard $R$ matrix and the latent $\boldsymbol{t_i}$ vector in the model is not learnt but kept static by setting each element in the vector to 1.
	
\end{itemize}

\par
\begin{figure}[t]
	\begin{minipage}[b]{\linewidth}
		\centering
		\includegraphics[width=\textwidth, height=6cm]{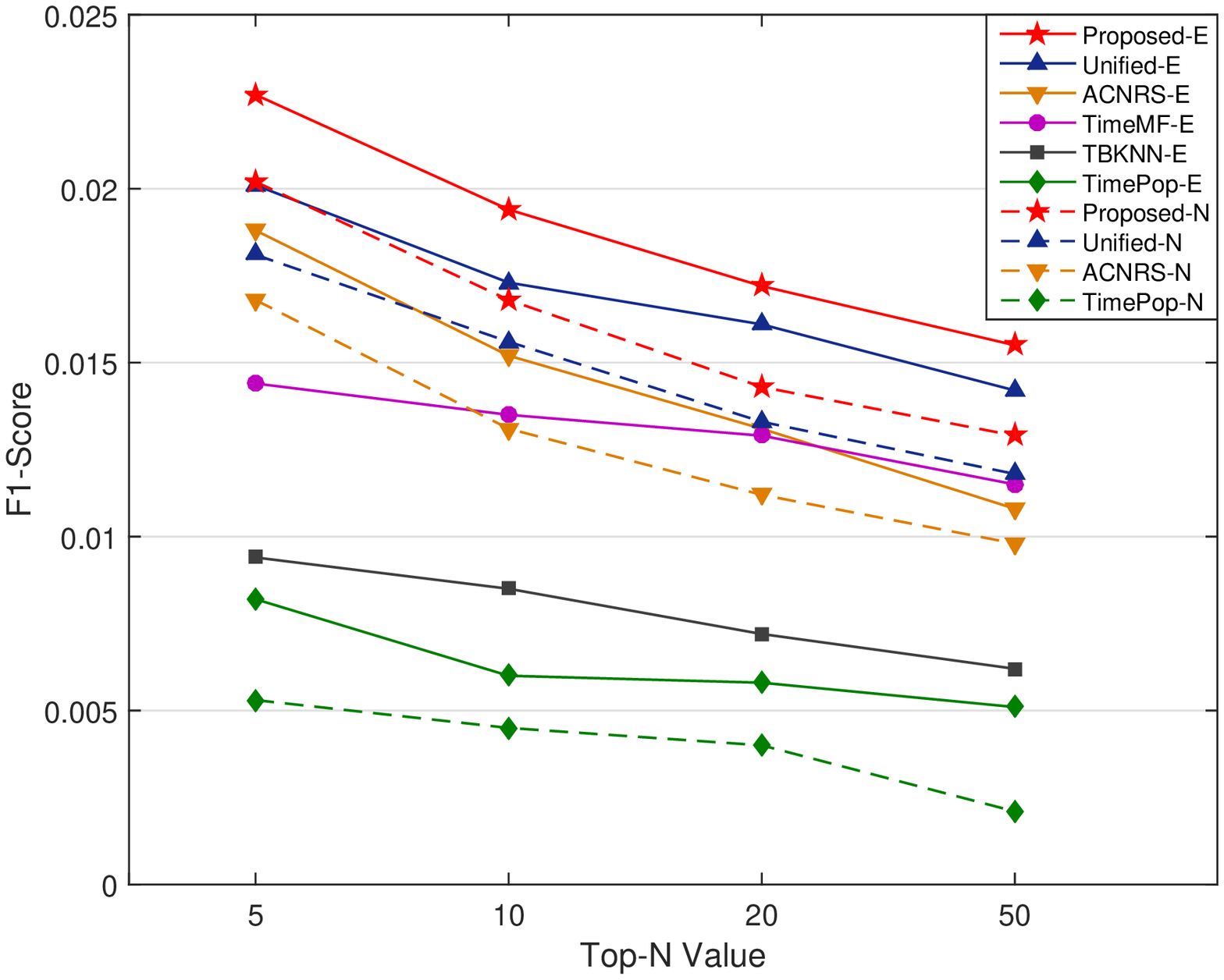}
		\caption{Results of Exp. 1 (Biweekly intervals with a training/testing ratio of 83/17).}
		\label{fig:pics5_results_w_83}
	\end{minipage}
	\hspace{0.5cm}
	\begin{minipage}[b]{\linewidth}
		\centering
		\includegraphics[width=\textwidth, height=6cm]{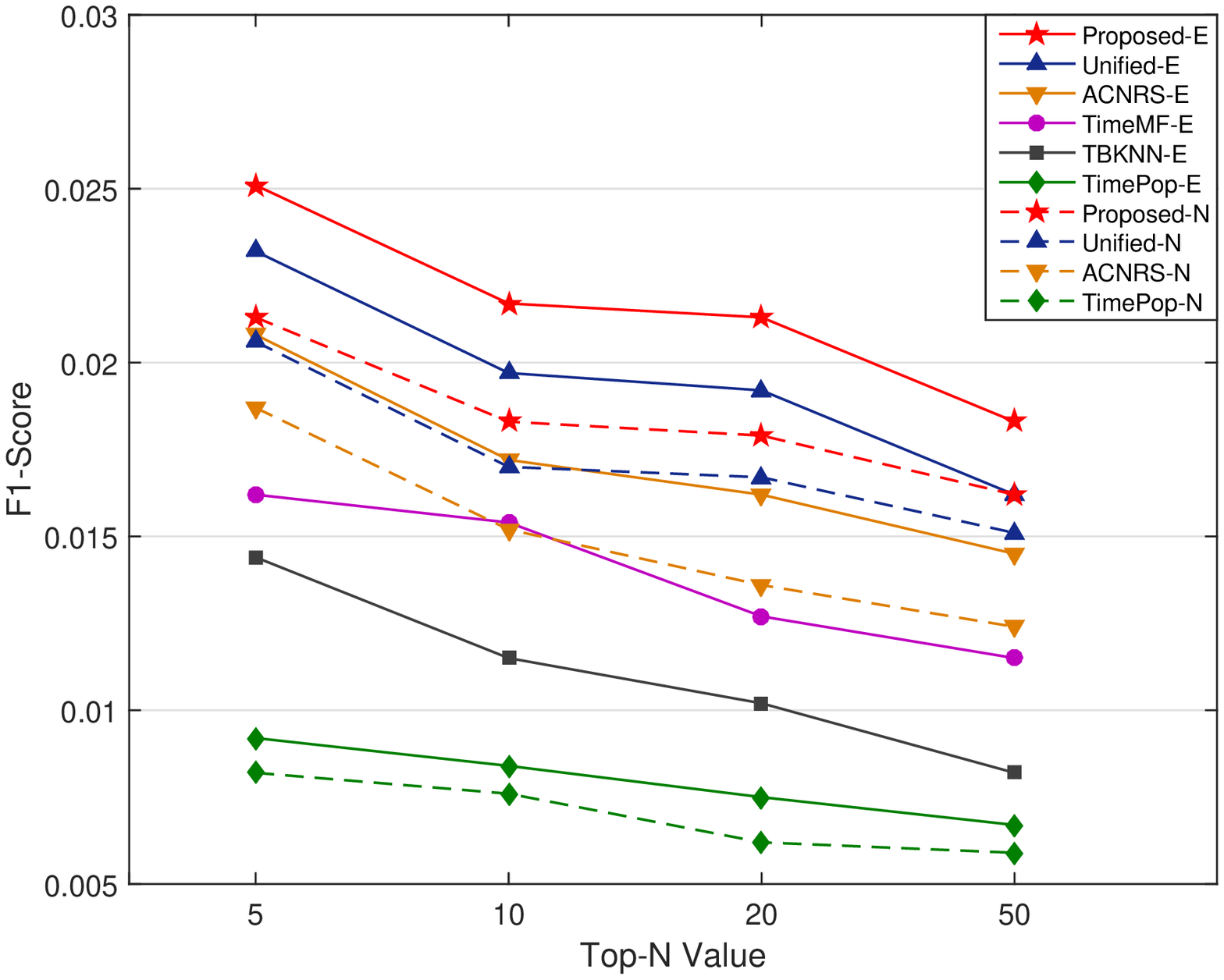}
		\caption{Results of Exp. 2 (Biweekly intervals with a training/testing ratio of 92/8).}
		\label{fig:pics6_results_w_92}
	\end{minipage}
\end{figure}
\begin{figure}[t]
	\begin{minipage}[b]{\linewidth}
		\centering
		\includegraphics[width=\textwidth, height=6cm]{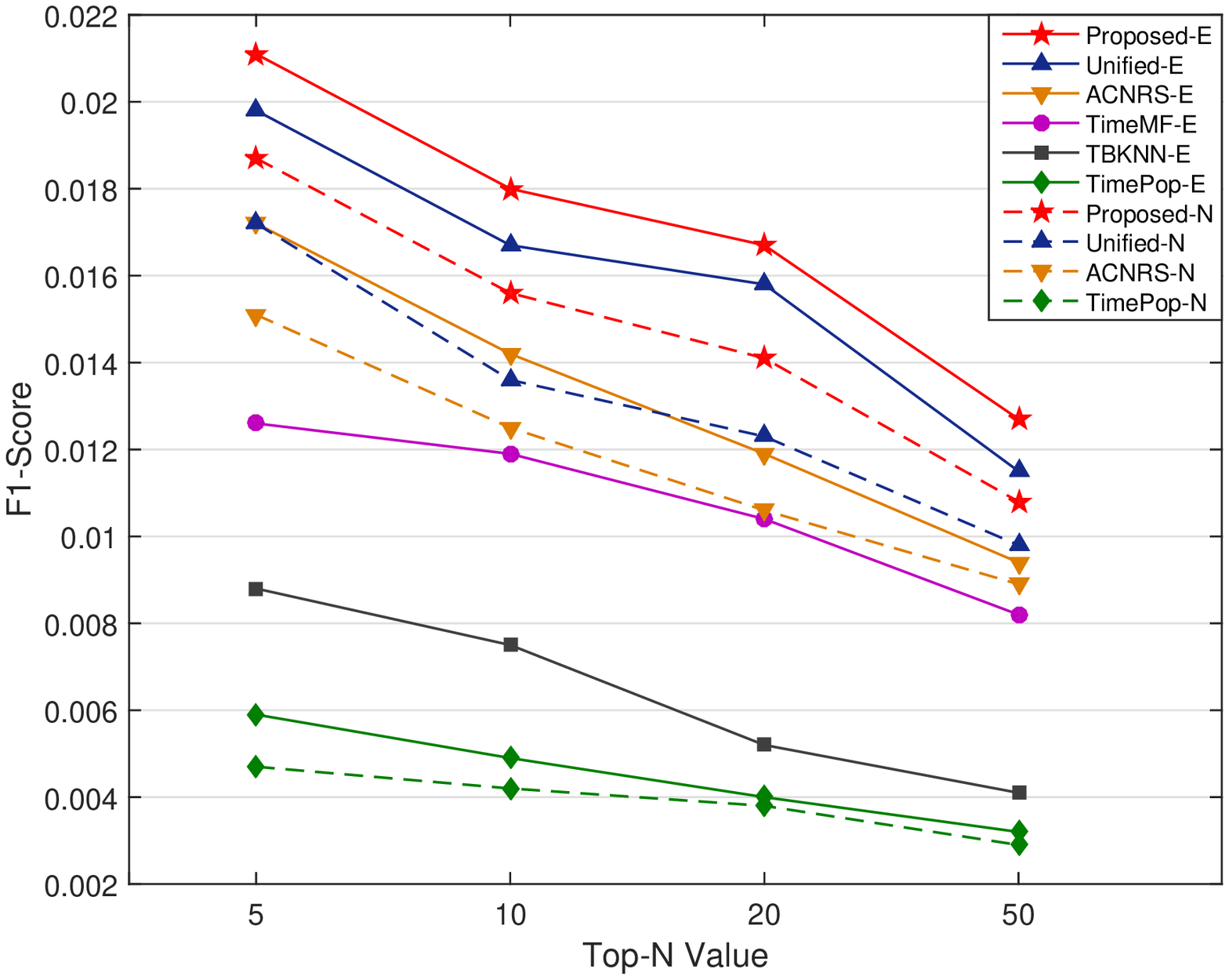}
		\caption{Results of Exp. 3 (Monthly intervals with a training/testing ratio of 83/17).}
		\label{fig:pics7_results_m_83}
	\end{minipage}
	\hspace{0.5cm}
	\begin{minipage}[b]{\linewidth}
		\centering
		\includegraphics[width=\textwidth, height=6cm]{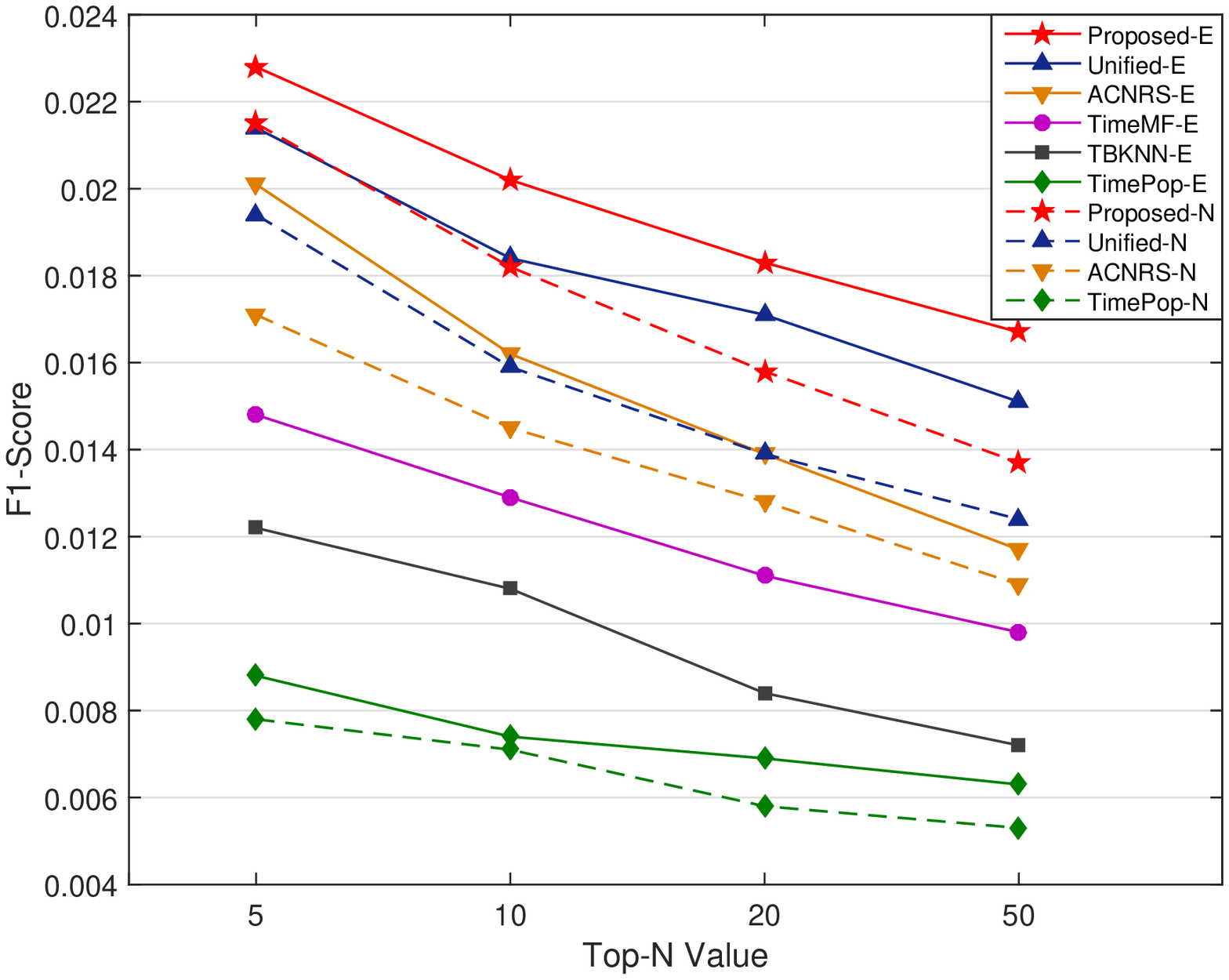}
		\caption{Results of Exp. 4 (Monthly intervals with a training/testing ratio of 92/8).}
		\label{fig:pics8_results_m_92}
	\end{minipage}
\end{figure}

In line with common practices in the literature, we  utilized popularity based and MF based baselines for model evaluation. Moreover, both time aware and cross-network methods were used for effective comparisons. The TimePop, Unified and ACNRS methods conduct recommendations for both new and existing users. The absence of training data for new users limits the predictions from TBKNN and TimeMF methods to only existing users. 
Overall, we examined 6 models in 4 experimental setups, for both new and existing user groups, which resulted in 40 result sets (see Figures \ref{fig:pics5_results_w_83} - \ref{fig:pics8_results_m_92}).
\par


Video recommendation is treated as a top-N recommendation task (see Section \ref{modpre}). Therefore, to measure model accuracy, we calculated F1-scores for the selected set of N predictions (5, 10, 15 and 20) as follows:
\begin{equation}F-Score@N = 2 \times \frac{Precision@N \times Recall@N}{Precision@N + Recall@N}\end{equation}

However, obtaining a higher recommendation accuracy alone is insufficient to measure broader aspects of user satisfaction \cite{mclaughlin2004collaborative}. For example, recommending a set of similar items (i.e., in terms of topic, author, genre, etc.) which represent a small fraction of user\textquoteright s preferences would result in higher accuracy, but is less useful and less interesting to the user. Therefore, we calculated two other metrics (i.e., diversity and novelty of recommended items) to broadly evaluate the effectiveness of the proposed solution. 
To measure the diversity between videos in the recommended item list ($Rec_l$), we utilized the following measure introduced by Avazpour et al. \cite{avazpour2014dimensions}: 
\begin{equation}Diversity(Rec_l) = \frac{\Sigma_{v_i \in Rec_l} \Sigma_{v_j \in Rec_l, v_i \neq v_j} (1-sim(v_i, v_j))}{s(s-1)/2}\end{equation}  
where, $s$ is the number of videos in the list and $Sim(v_i, v_j)$  is a normalized similarity measure of $v_i$ and $v_j$ videos, obtained using the intersection of their topical distributions as follows:
\begin{equation}Sim(v_i, v_j) = \frac{\Sigma_{k=1}^{k^t}min(v_i^k, v_j^k)}{\Sigma_{k=1}^{k^t}(v_i^k + v_j^k)}\end{equation}  
where, $v_i^k$ denotes the frequency of the $k^{th}$ topic of video $v_i$. 
\par
The novelty metric is used to measure the extent of unfamiliarity between successfully recommended items. We utilized a novelty measure similar to the measure used by Zhang \cite{zhang2013definition} as follows:
\begin{equation}Novelty(Rec_l) = \frac{\lvert Rec_l \cap Tes_l \lvert}{\lvert Tes_l \lvert} - \frac{\lvert Rec_l \cap Tra_l \lvert}{\lvert Tra_l \lvert}\end{equation}
where, $Tes_l$ and $Tra_l$ are videos in the testing and training sets. The novelty value rewards the system for successfully recommended videos and penalizes for recommending previously interacted items.

\subsection{Model Parameters}
\label{modpar}
We set the following model parameter values, which were experimentally justified. 
\begin{itemize}
	\setlength\itemsep{0em}
	\item 
	When using Twitter-LDA to project user interactions to a cross-network topical space, we set the number of topics, $K^t$ to 60 using a grid search algorithm. The final results were not highly sensitive to minor changes in $K^t$ $(\approx \pm 5)$.  
	\item 
	We used a grid search algorithm and a two-fold cross validation setup to obtain $K$ and $\lambda$ values, which were set to 60 and 0.5, respectively. 
	\item 
	By configuring the model using the above parameters and using a grid search algorithm, we found optimum values for $\beta$ and $\gamma$ to be 0.8 and 0.3, respectively. 
	\item
	The learning rate $\mu$ was set to a fairly small value, 0.001, to obtain the local minimum.
\end{itemize}
\subsection{Discussion}
\label{resuts}
The effectiveness of the proposed model is validated by comparing against other baseline methods under different experimental setups (see Table \ref{tab:expsett}). We summarized the average F1-score values for each user type (new and existing) under varying top-N conditions (5, 10, 20 and 50) (see Figures \ref{fig:pics5_results_w_83} - \ref{fig:pics8_results_m_92}) along with novelty and diversity measures (see Table \ref{tab:divnow}). Note that in figure legends, the suffixes -N and -E refer to new and existing user recommendations, respectively.
\par
It can be seen that in general, accuracy values for biweekly intervals (see Figures \ref{fig:pics5_results_w_83}, \ref{fig:pics6_results_w_92}) are comparatively higher than monthly intervals (see Figures \ref{fig:pics7_results_m_83}, \ref{fig:pics8_results_m_92}). This is intuitive because the smaller time gaps allow user profiles to be modeled using finer level user preference dynamics. Furthermore, under both time granularities, a larger training set (Exp. 2 and Exp. 4) leads to marginally higher results, possibly due to better trained models. As expected, among various Top-N values, the highest accuracy is achieved by the Top-5 recommendations. The accuracy generally declines with higher Top-N values, due to increasing false positive rates. 
\par
Note that both TBKNN and TimeMF models cannot provide recommendations for new users because these models can only be trained using rating values from the target network (i.e., single network based). The absence of new users\/' ratings makes these models inoperable for new user predictions. Unlike other single network based solutions, TimePop is able to conduct both new and existing recommendations. However, comparatively, TimePop achieves relatively low accuracy  levels. This is expected as it does not provide personalized recommendations. Comparatively, all three cross-network solutions (ACNRS, Unified and Proposed) are most effective for both new and existing user recommendations. 
\par 
Furthermore, the increase in accuracy when incorporating timing information for user recommendations is evident. Compared to the time unaware alternative model (ACNRS), the proposed model achieves notable accuracy improvements under all experimental setups.
Moreover, across all top-N conditions, the proposed model achieves an average of 9.4\% accuracy increase compared to the considerable time unaware cross-network (Unified) approach. 
\par
Additionally, compared to other methods, the proposed model shows better novelty and diversity among recommended items. On average, the proposed model shows a 2.2\% and 11.8\% improvement in novelty and diversity measures over the closest Unified approach (see Table \ref{tab:divnow}). Hence, it is evident that the proposed time aware cross-network model considerably improves the overall recommender quality.
\begin{table}[h!]
	\small
	\centering
	\caption{A comparison of Novelty and Diversity measures.}
	\label{tab:divnow}
	\begin{tabular}{|l|l|l|l|l|}
		\hline
		Metrices  & TimePop & TimeMF & Unified & Proposed \\ \hline
		Diversity & 0.6823  & 0.6686 & 0.7265  & 0.7423   \\ \hline
		Novelty   & 0.0159  & 0.0151 & 0.0187  & 0.0209   \\ \hline
	\end{tabular}
\end{table}

\section{Conclusions and Further work}
\label{concfur}
Recent efforts in both cross-network and time aware recommender systems show considerable improvements over standard systems. However, to the best of our knowledge, this is the first effort to combine both fields to successfully address two coexisting issues in recommender systems: (1) incomplete user profiles and (2) the dynamic nature of user preferences. Hence, we developed a recommender solution that, first, transfers historical time aware user preferences from both source and target networks to a target network user space and second, aggregates them to obtain the current user preference models for timely recommendations in the target network. To validate the proposed method, we proved its effectiveness against five baseline methods (three time based and two cross-network based) using accuracy, diversity and novelty measures under various experimental settings. The experimental results showcased that addressing a single limitation bounds the achievable recommender quality and the proposed time aware cross-network recommender solution achieves superior recommender quality by addressing both coexisting limitations in a single solution. 
\par
We identify the following limitations in our proposed solution, which could be addressed as future work: (1) only uses textual data from Twitter and YouTube to create user profiles, which could be extended to incorporate other data sources and multi-modal data (e.g., images and videos). (2) models linear relationships among factors, which could be extended to non-linear models (e.g., non-linear MF and deep learning based models) that better capture underlying complex relationships to further improve the recommender quality.
\par
The recommender systems literature showed that time aware information can be used to improve the recommender quality in single network based solutions. Our solution showcased that the use of time aware information can improve the quality of cross-network recommender systems. Therefore, we believe that these results lay a foundation for future work to exploit time aware cross-network systems to obtain better recommender quality.
	
	\subsection*{Acknowledgments}
	This research was supported in part by the National Natural Science Foundation
	of China under Grant no.~61472266 and by the National University of Singapore
	(Suzhou) Research Institute, 377 Lin Quan Street, Suzhou Industrial Park, Jiang
	Su, People's Republic of China, 215123.

	\bibliographystyle{ACM-Reference-Format}
	\bibliography{sigproc2} 
	
\end{document}